\documentclass[10pt, conference]{IEEEtran}

\ifCLASSINFOpdf
  \usepackage[pdftex]{graphicx}
\else
\fi
\begin{document}
\title{On Using Active Learning and Self-Training when Mining Performance Discussions on Stack Overflow}


\author{
\IEEEauthorblockN{Markus Borg}
\IEEEauthorblockA{Software and Systems Engineering Lab.\\
RISE SICS AB\\
Lund, Sweden\\
markus.borg@ri.se}
\and
\IEEEauthorblockN{Iben Lennerstad}
\IEEEauthorblockA{Dept. of Computer Science\\
Lund University\\
Lund, Sweden\\
iben.lennerstad@gmail.com}
\and
\IEEEauthorblockN{Rasmus Ros, Elizabeth Bjarnason}
\IEEEauthorblockA{Dept. of Computer Science\\
Lund University\\
Lund, Sweden\\
firstname.lastname@cs.lth.se}
}

\IEEEspecialpapernotice{(Preprint of paper accepted for the Proc. of the 21st International Conference on Evaluation and Assessment in Software Engineering, 2017)}

\maketitle

\begin{abstract}
Abundant data is the key to successful machine learning.
However, supervised learning requires annotated data that are often hard to obtain. 
In a classification task with limited resources, Active Learning (AL) promises to guide annotators to examples that bring the most value for a classifier.
AL can be successfully combined with self-training, i.e., extending a training set with the unlabelled examples for which a classifier is the most certain.
We report our experiences on using AL in a systematic manner to train an SVM classifier for Stack Overflow posts discussing performance of software components. 
We show that the training examples deemed as the most valuable to the classifier are also the most difficult for humans to annotate.
Despite carefully evolved annotation criteria, we report low inter-rater  agreement, but we also propose mitigation strategies.
Finally, based on one annotator's work, we show that self-training can improve the classification accuracy.
We conclude the paper by discussing implication for future text miners aspiring to use AL and self-training.
\end{abstract}


\section{Introduction}
% PROVIDE AN ACCOUNT OF THE NEW RESEARCH. MAKE CLEAR IT IS NOVEL AND SIGNIFICANT.
%
% Establish a territory move (link the problem studied to the general research area)
%	- Claim centrality step (this is important to us all because...)
%	- Indicate a territorial lack or problem step
%	- Present background/topic generalizations step
%	- Review previous research step
%
% Establish a niche move (To indicate there is a need within the research area)
%	- Indicate lacks in previous research step
%	- Indicate that the present research follows a tradition step
%	- Indicate an unsolved research or real-world problem
%	- Make counterclaims step (very strong to do)
%
% Occupy the niche move
%	- Announce objectives step
%	- Announce principal findings step
%	- Announce or briefly describe the scope of the present research step
%	- Refer to important aspects step
%	- Announce the RA structure step

Large datasets are key to successful machine learning and text mining. 
For example, applying natural language related machine learning to text at web scale~\cite{pereira_unreasonable_2009} has enabled many of the advances in the last decade.
It is well known that an algorithm that works well on small datasets might be beaten by simpler alternatives as more data are used for training~\cite{banko_scaling_2001}.
However, while the web contains huge amounts of text, supervised learning requires annotated data -- data that are hard to obtain.

A common solution to acquire enough annotated data is crowdsourcing using services such as Amazon Mechanical Turk.
The possibility to employ a massive, distributed, anonymous crowd of individuals to perform general human-intelligence micro-tasks for micro-payments has radically changed the way many researchers work~\cite{mitra_comparing_2015}.
However, when annotation requires more than general human intelligence, i.e., for non-trivial micro-tasks, such crowdsourcing solutions might not work.
Annotation of developers' posts on Stack Overflow is an example of non-trivial classification for which successful crowdsourcing cannot be expected. 
%A large enough set of experienced annotators is often difficult to find, despite screening of workers and providing detailed instructions. 

Active Learning (AL) is a semi-automated approach to establish a training set. 
The idea is to reduce the overall human effort by focusing on annotating examples that maximize the gained learning, i.e., the examples for which the classifier is the most uncertain. 
AL has been used for software fault prediction, successfully reducing the need for human intervention~\cite{sun_carial:_2012,lu_adaptive_2012,lu_defect_2014}.
%We have not found any other examples from the MSR community that use crowd sourced information for AL.
AL has also been used in several other fields of research, e.g., for creating large training sets for speech recognition and information extraction~\cite{settles_active_2010}.
Several studies show that AL can successfully be combined with self-training, which is a method to extend the training set by automatic labeling of a trained classifier~\cite{lin_combining_2010,zhang_active_2016,richarz_annotating_2012}, but the techniques have not previously been used for text mining Stack Overflow.

In this study, our target training set is \textit{Stack Overflow discussions on performance of software components}.
Our work is part of the ORION project, in which we aim at developing a decision-support system for software component selection~\cite{wohlin_supporting_2016}. 
One aspect under study is how to collect and store experiences from previous decisions~\cite{cicchetti_towards_2016}.
The ORION project proposes collecting experiences from both internal and external sources, i.e., both from the company and from other organizations.
In this paper, we address using machine learning to extract external experiences from the software engineering community by text mining Stack Overflow, the leading technical Q\&A platform for software developers~\cite{barua_what_2012}.

We report our experiences from using AL and an SVM classifier in a systematic way consisting of 16 iterations.
Our findings show that not only the classifier is uncertain regarding the borderline cases -- also the human annotators display limited agreement.
Consequently, we stress that annotation criteria must continuously evolve during AL.
Moreover, we suggest that AL with multiple annotators should be designed with partly overlapping iterations to enable detection of different interpretations.  
Finally, we demonstrate that self-training has the potential to improve classification accuracy.

The rest of the paper is organized as follows: Section~\ref{sec:back} introduces background and related work, Section~\ref{sec:method} presents the design of our study, and Section~\ref{sec:results} discusses our findings. 
Finally, we summarize our implications for future mining operations in Section~\ref{sec:conc}.

\section{Background and Related Work} \label{sec:back}
\begin{figure}
\centering
\includegraphics[width=0.5\textwidth]{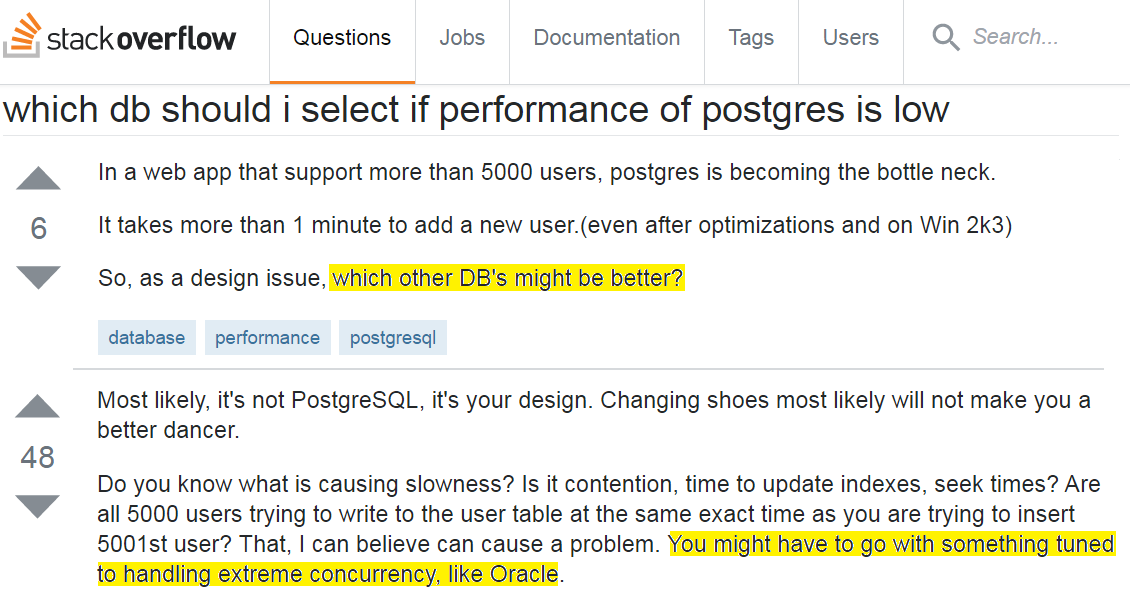}
\caption{Example of a performance question with an answer on Stack Overflow. We highlight the parts of the text particularly relevant. The figure also shows the question's three tags.}
\label{fig:example}
\end{figure}

Stack Overflow is the dominant technical Q\&A platform for software developers, with 101 million monthly unique visitors (March 2017).
The information available on Stack Overflow has been studied extensively in the software engineering community, mostly through text mining, but also through qualitative analysis.
Fig.~\ref{fig:example} shows an example of a Stack Overflow question with an answer, in which we highlight text chunks related to performance.

Treude \textit{et al.} investigated the type of questions asked and the quality of the answers and found that the information is particularly useful for code reviews and conceptual questions, and for novice developers ~\cite{treude_how_2011}. 
Soliman \textit{et al.} found that Stack Overflow contains information relevant to and useful for decisions within software architectural design, and have idenitified a list of words that may be used to automatically classify such information ~\cite{soliman_architectural_2016}. 
Topic modelling has been used to identify what topics that are discussed and relationships between these. 
In this way, Barua \textit{et al.} identify a number of current trends within software development, e.g., that mobile app development is increasing faster than web development ~\cite{barua_what_2012}. 
It is suggested that knowledge mined from Stack Overflow can be used to provide context-relevant hints in IDEs ~\cite{allamanis_why_2013,ponzanelli_mining_2014} and for filtering out off-topic posts, e.g., in chat channels ~\cite{chowdhury_mining_2015}.

\begin{figure}
\centering
\includegraphics[width=0.345\textwidth]{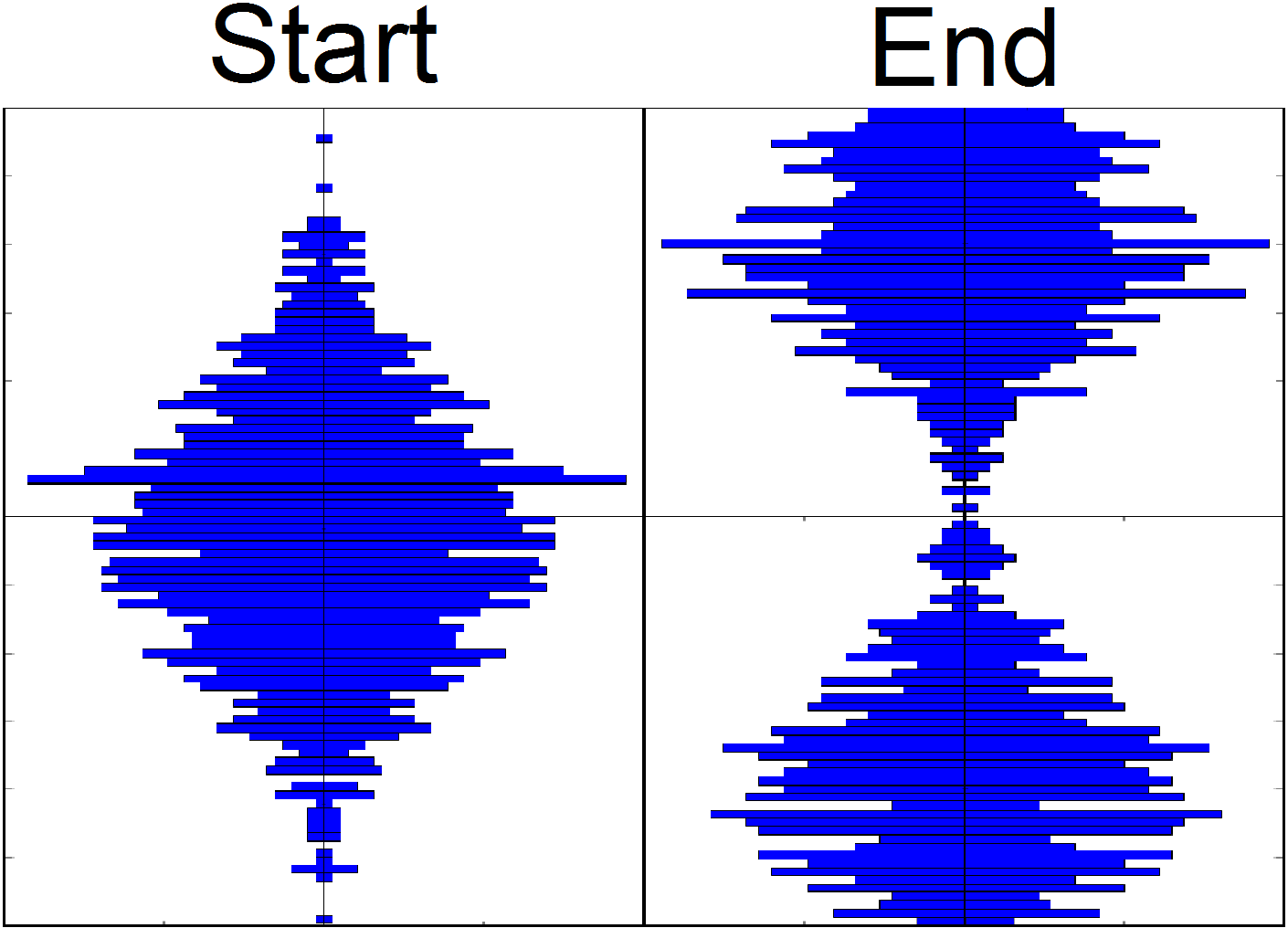}
\caption{Target effect on unlabelled examples when applying AL and SVM. To the left, many examples appear around the hyperplane (the horizontal line). To the right, after completing the AL iterations, the remaining unlabelled examples are separated, i.e., fewer borderline cases remain.}
\label{fig:AL}
\end{figure}

AL is a semi-supervised machine learning approach in which a learning algorithm interactively queries the human to obtain labels for specific examples, typically the most difficult ones.
The method for selecting examples to query should be optimized to maximize the gained learning. 
Uncertainty sampling is a simple technique that selects examples where the classifier is least certain on which label to apply~\cite{settles_active_2010}. 
This has the effect of separating the examples into two distinct groups and thus remove borderline cases, see the horizontal histograms in Fig.~\ref{fig:AL}.
AL enables a shift of focus from momentary data analysis to a process with a feedback loop~\cite{hassan_software_2010,menzies_inductive_2011}.

When mining from crowdsourced data there are usually too many unlabelled examples to annotate them all manually. 
Semi-supervised learning are methods that use also remaining unlabelled examples to improve the classifier. 
Self-training (or bootstrap learning~\cite{yarowsky_unsupervised_1995}) is one such method that extends the training set with the unlabelled examples classified with the highest degree of certainty.
This complements AL with uncertainty sampling well, since it maximizes the available confident labels~\cite{settles_active_2010}.
To the best of our knowledge, we present the first application of both AL and self-training for Stack Overflow mining.

\section{Method} \label{sec:method}

\begin{figure}
\centering
\includegraphics[width=0.5\textwidth]{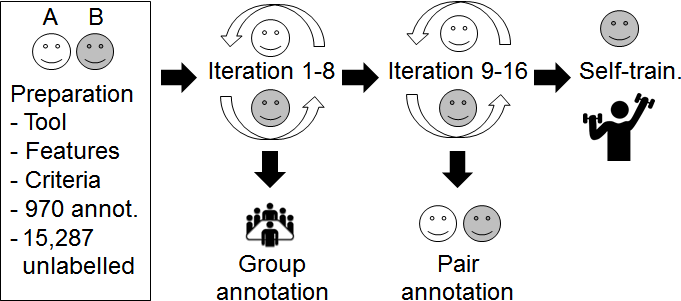}
\caption{Overview of the research design. Smileys depict where in the process the first author (A) and the second author (B) where involved. Note that for the self-training step, the smiley shows that the second author's annotated data was used in the automatic process.}
\label{fig:design}
\end{figure}

We designed a study to evaluate AL when mining Stack Overflow.
Fig.~\ref{fig:design} shows an overview of the research design that consisted of a preparation step and two iterative training steps.
In the preparation step, we downloaded the dataset used for the MSR Mining Challenge in 2015 containing 43,336,603 posts~\cite{ying_mining_2015}.
We extracted all that were tagged with `performance' and at least one of the following tags: `apache', `nginx' or `rails' -- an attempt to get an initial dataset related to components we know well, resulting in 2,304 posts in total.

\textbf{Preparation} To assist the manual annotation task, we developed a prototype tool integrating an SVM classifier from scikit-learn~\cite{pedregosa_scikit-learn:_2011}, i.e., the classifier finds the optimal \textit{hyperplane} separating two categories of examples~\cite{vapnik_nature_2000}. 
In our application, we trained an SVM classifier with n-grams as features (n=1-5) to separate Stack Overflow posts related to performance discussions of software components and other posts.
We refer to the two categories as positive and negative examples, respectively.
%The tool displays the unlabelled post closest to the SVM hyperplane, allowing the user to mark it as positive before before clicking the `Next' button.
%Also, the user highlight component names to add them to a separated list (described later in this section).

During the tool development, the first and second authors alternated annotating posts and evolving initial annotation criteria -- note that this inital step was done without AL.
In total, we annotated 970 posts (25.4\% positive) and the criteria evolved into ``a positive post discusses the performance of a software component, rather than programming languages, the development environment, or measurements tools''. 
While manually annotating the initially posts, we identified 67 additional component names that also had explicit Stack Overflow tags.
We used this to extend our dataset, i.e., we complemented `apache', `nginx' or `rails' with 67 new tags to obtain a larger dataset of Stack Overflow posts.
In total we collected 15,287 Stack Overflow posts potentially related to performance of software components\footnote{Replication package: URL}.

\textbf{Active learning} After the preparation, the first and second authors alternated manual annotation of the next 100 posts\footnote{A reasonable annotation task that requires roughly 90 min.} closest to the SVM hyperplane --
we refer to each such annotation batch~\cite{settles_theories_2011} as an AL \textit{iteration}.
For each iteration, we measured the classification accuracy complemented by precision, recall, and $F_{1}$-score using 5-fold cross-validation.
Furthermore, we calculated the distance from each post, both labelled and unlabelled, to the SVM hyperplane.
We visualize the distribution of posts at different distances from the SVM hyperplane using histograms and beanplots.

\textbf{Self-training} We investigated self-training based on the second author's annotation activity (cf. `Self-train. in Fig.~\ref{fig:design}) by adding unlabelled examples as if they were manually annotated.
We explored extending the training set with different percentages of unlabelled data, corresponding to different distances to the SVM hyperplane. 
Our ambition was to identify a successful application of self-training, useful as a proof-of-concept, rather than finding the optimal parameter settings for this particular case.

\textbf{Human annotation} To measure the uncertainty in classifying Stack Overflow posts close to the SVM hyperplane, we evaluated the inter-rater reliability of human annotators.
The first and second author discussed experiences after each completed iteration, and the annotation criteria evolved.
After 8 iterations, halfway into the study, we considered the criteria mature enough for evaluation. The criteria were then:

``A positive post (both questions and answers) \textit{addresses the performance of a specific software component} (incl. frameworks, platforms, and libraries) that could be used to evolve a software-intensive system.
Examples: database management systems (MySQL, Oracle, ..), content management systems (Drupal, Joomla, ..), web servers.

A post is negative if it discusses performance of/from:
\begin{itemize}
\item programming languages (e.g., Java, PHP)
\item operational environments (e.g., Windows, Linux)
\item development tools (e.g., compilers, IDEs, build systems.)
\item alternative detailed implementations (e.g., formulation of SQL queries, parsing of XML/JSON structures)  
\item tweaking of components
\end{itemize}
or if the post discusses components used to measure performance (e.g., JMeter, SQLTest).
The exclusion criteria apply, \textit{unless such a discussion clearly originates in poor performance of a specific component}''.

We designed a hands-on annotation exercise during a research workshop with 12 senior software engineering researchers (cf. `Group annotation' in Fig.~\ref{fig:design}).
First, we introduced the exercise, showed some examples, and provided the above criteria.
Second, everyone independently annotated 11 posts, printed on paper, during a 20 minute session.
In total, 66 posts were distributed using pairwise assignment: two annotators per post, and each possible human pair represented once.
Finally, we calculated Krippendorff's $\alpha$ to assess inter-rater reliability, as recommended for difficult nominal tasks~\cite{feng_mistakes_2015}.

After the group annotation, we discussed the outcome to better understand our differences.
%The first and second authors hypothesized that by annotating more posts, the shared understanding would grow.
We continued the annotation activity, following the same process and expecting a growing shared understanding, until iteration 16.
Once finished, we had 2,567 annotated posts (32.6\% positive).
To check our hypothesis of improving agreement, we randomly selected 50 posts among the already annotated (cf. `Pair annotation' in Fig.~\ref{fig:design}).
Again we calculated Krippendorff's $\alpha$, both 1) between the first and second authors (referred to as A and B), and 2) between the new labels and the previous labels.
For each post annotated differently, we quantified the certainty of the set label (1-5) and we provided a rationale.

\section{Results and Lessons Learned} \label{sec:results}
% PRESENT THE NOVELTY
% Move 1: Create a context for the novelty claim
%	- Step 1: Five a brief overview of the main points covered in the R-section
%	- Step 2: Provide background to the audience to support appreciation of results
%	- Step 3: Explain the need for the results
%	- Step 4: Provide information about the organization of the section
%	- Step 5: Summarize aspects of the methods that are important to understand results (if reader skips M-section)
%
% Move 2: Establish novelty
%	- Step 1: Refer to facts that show the main novel outcomes
%	- Step 2: Refer to facts that show the main type of novel results
%	- Step 3: Refer to important new relationships discovered
%	- Step 4: Refer to the innovations/modifications/applications/viability of the outcomes
%
% Move 3: Strengthen the novelty claim
%	- Step 1: Point out that data (usually in figures) confirm novelty
%	- Step 2: Demonstrate that the data confirm novelty or validity
%	- Step 3: Explain important or non-obvious relationships between outcomes
%	- Step 4: Indicate that the findings are valid/reliable/accurate/useful etc.
%	- Step 5: comment on the interest or importance of the outcomes

%\subsection{Active Learning and Stack Overflow (RQ1)}
%Our primary goal was to train an accurate classifier, thus we incrementally created learning curves illustrating the effects of growing the training set adding new iterations.
\textbf{Human annotation} We begin this section by reporting on the inter-rater reliability.
The results from our group annotation exercise after 8 iterations confirmed the challenge of annotating posts close to the SVM hyperplane.
Despite annotation criteria that evolved during 8 AL-iterations, the 12 annotators obtained a Krippendorff's $\alpha$ of 0.126 (37/66 shared labels, 56\%) -- a poor agreement.
The first and second authors analyzed the discrepancies, along with posts for which there were agreement, without identifying any concrete patterns.  
The presence of borderline cases is obvious, but we hypothesized that the alignment between the first and second authors was stronger than within the whole group, and that it would continue improving during the remaining iterations.

After 16 AL-iterations, we calculated the inter-rater reliability between  A and B for a random sample of 50 previously annotated posts.
The exercise yielded a Krippendorff's $\alpha$ of 0.028 (29/50 shared labels, 58\%), considerably lower than from the group exercise.
We also calculated the inter-rater reliability against our previous annotations of the 50 posts, obtaining a Krippendorff's $\alpha$ of 0.768 (18/20 shared labels, 90\%), and 0.577 (24/30 shared labels, 80\%) for A and B, respectively.
Our results show that while our individual annotation remained stable over time, our shared view still differed after 16 iterations. 

In most cases at least one of us was very uncertain, expressing a certainty level of 1 or 2, which means the post was more or less randomly labelled.
More alarming, however, was that in several cases both annotators felt certain but used different labels.
An analysis of the latter cases revealed that A was more inclusive regarding posts that related to implementation details and component tweaking, whereas B was more inclusive concerning quality attributes not necessarily related to performance.
Furthermore, B did not include posts that could be interpreted as anecdotal experiences.
We conclude that AL for text classification is difficult, even after annotating 2,674 posts with several intermediate discussions, our inter-rater reliability was low.

\textbf{Active learning} Since our annotation criteria did not properly align our annotation activity, we hesitated to pool our training data.
Instead, we trained three separate SVM classifiers using: 1) A data, 2) B data, and 3) A+B data -- we refer to these as SVM A, SVM B, and SVM A+B, respectively.
Note that we also split the training data from iteration 0 into either A or B, resulting in differently large initial training sets.

\begin{figure}
\centering
\includegraphics[width=0.5\textwidth]{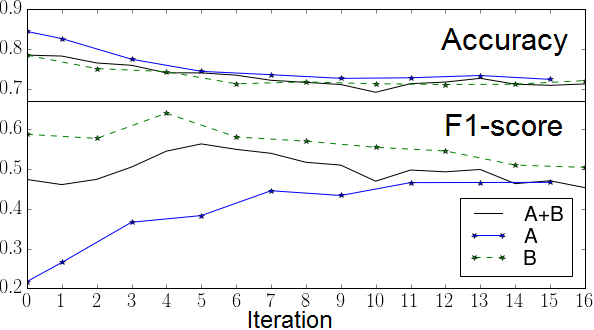}
\caption{Learning curves for the SVM classifier. Note that the 0th iteration contained 373 labelled examples for A, and 600 for B. Each subsequent iteration adds another 100 examples.}
\label{fig:learningCurves}
\end{figure}

Fig.~\ref{fig:learningCurves} shows the mean value from five runs of 5-fold cross-validation for each iteration.
The solid lines with markers show accuracy and $F_{1}$-score for SVM A, the dashed lines with markers represent SVM B, and the solid lines without markers illustrate SVM A+B.
Regarding accuracy, all three classifiers show similar behavior: 
The accuracy decreases as additional iterations are added, but the differences are minor.
The curves do not resemble typical learning curves, instead they appear to stabilize between 0.7 and 0.8. 
We explain this by the posts annotated for iteration 0, i.e., clearly positive and negative examples were selected to span the document space, followed by nothing but borderline cases selected using AL. 
Looking at $F_{1}$-score, SVM B and SVM A+B remain fairly stable around 0.5.
On the other hand, SVM A improves considerably as more iterations are added. 
This is likely due to the distribution of examples in the small A iteration 0 training set, containing only 373 examples and a recall of only 0.18 -- even adding borderline cases was useful in this case.

\begin{figure*}
\centering
\includegraphics[width=0.6\textwidth]{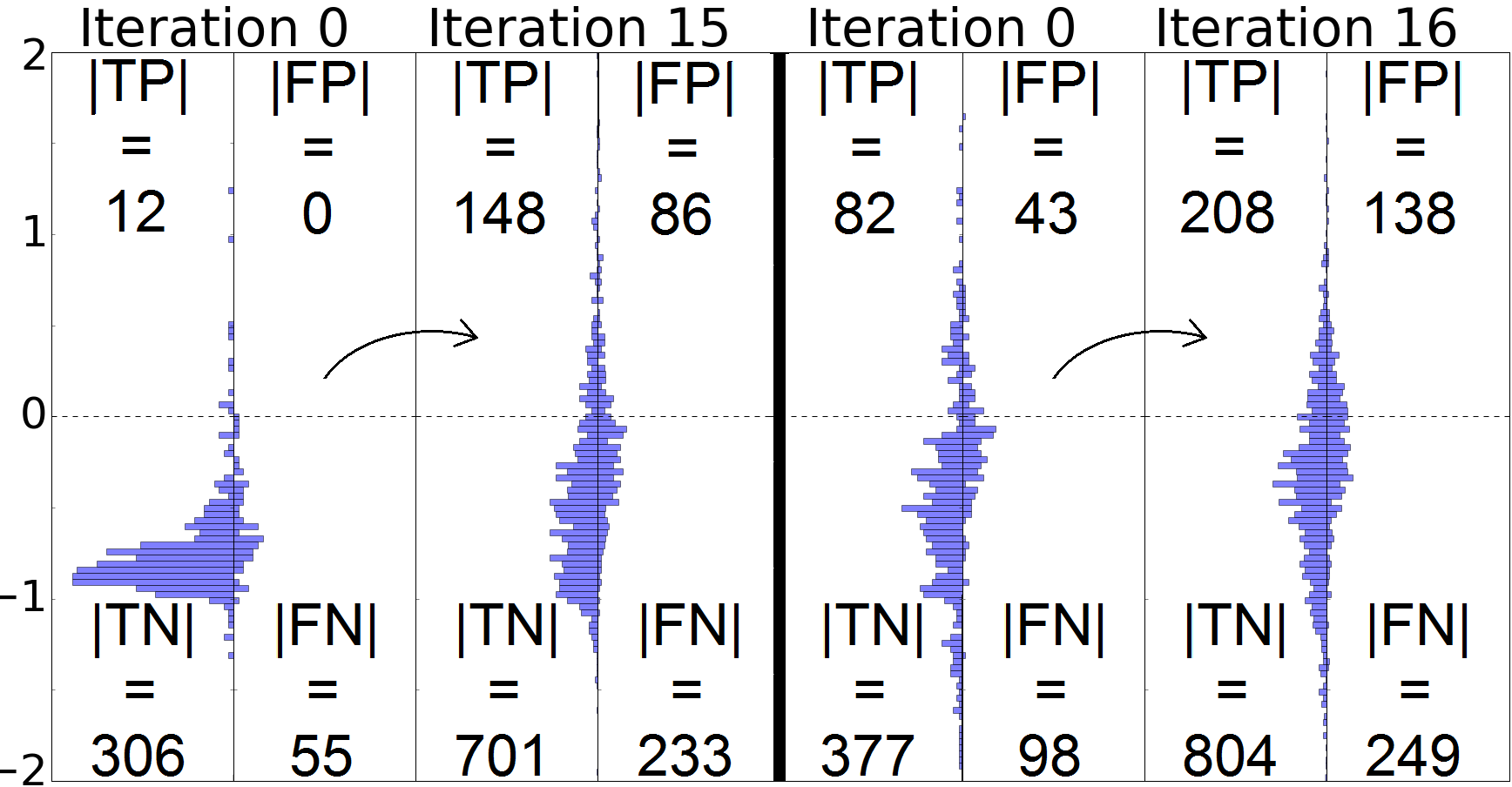}
\caption{Distribution of distances from labelled posts to SVM hyperplanes: SVM A iteration 0 and 15 (left), SVM B iteration 0 and 16 (right).}
\label{fig:beanLabelled}
\end{figure*}

Fig.~\ref{fig:beanLabelled} depicts distances between annotated posts and the SVM hyperplanes (SVM A and SVM B) after the preparation step and after the final iterations.
The vertical histograms show frequency distributions of posts with distances from the hyperplane on the y-axis, where the sign denotes positive and negative classifications, respectively.
Moreover, the figure displays the number of true positives (TP), false positives (FP), true negatives (TN), and false negatives (FN).
We notice that as more posts are annotated, the distribution around the hyperplane increases, which is particularly evident for the true negatives.
This shows that, from the perspective of the SVM classifiers, 16 AL iterations did not reduce the number of borderline posts.

% Annot. A Iter 0
% Pr = 0.850
% Rc = 0.176
% F1 = 0.292

% Annot. A Iter 15
% Pr = 0.636
% Rc = 0.393
% F1 = 0.486

% Annot. B Iter 0
% Pr = 0.629
% Rc = 0.367
% F1 = 0.464

% Annot. B Iter 16
% Pr = 0.625
% Rc = 0.410
% F1 = 0.495

\begin{figure*}
\centering
\includegraphics[width=0.55\textwidth]{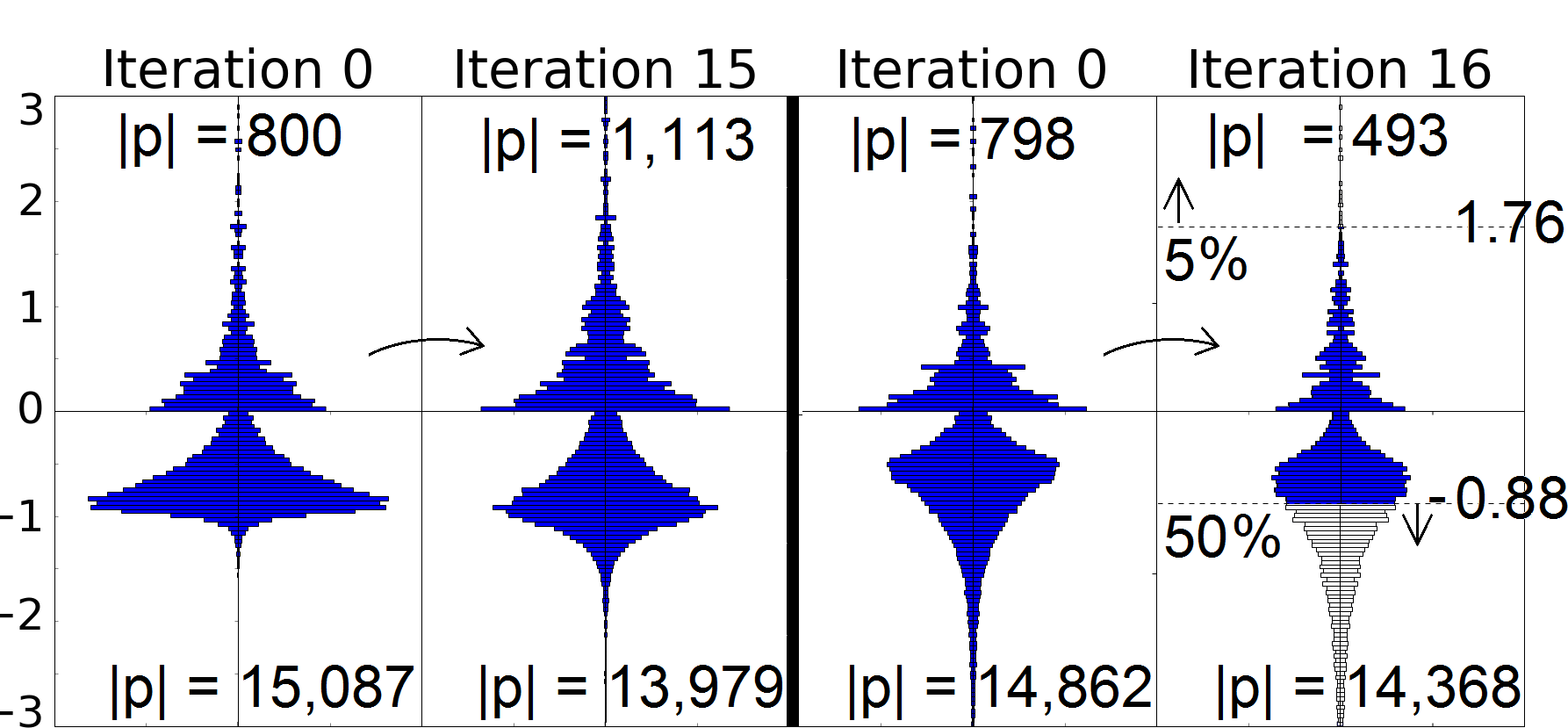}
\caption{Distribution of distances from unlabelled posts to SVM hyperplanes: SVM A iteration 0 and 15 (left), SVM B iteration 0 and 16 (right). Note that the scale on the negative side represents 10x as many posts as the positive side.}
\label{fig:beanUnlabelled}
\end{figure*}

Fig.~\ref{fig:beanUnlabelled} presents an analogous view for the unlabelled posts, also separating SVM A and SVM B.
However, in this figure we show beanplots, i.e., the frequencies are mirrored on the y-axis.
We also report the number of unlabelled posts on both sides of the hyperplanes (cf. $|p|$).
SVM A suggests that there are 716 positive posts remaining in the set of set 13,745 posts, whereas SVM B gives 259 remaining positive posts -- these figures reflect A's more inclusive interpretation of the annotation criteria.
The goal of AL is to focus annotation efforts on borderline cases to create two clearly separated clusters of examples (cf. Fig.~\ref{fig:AL}).
This phenomenon is not obvious in the Fig.~\ref{fig:beanUnlabelled}, although we observe that SVM B indeed has fewer negative examples close to the hyperplane, i.e., the beanplot close to 0 is thinner after iteration 16.
The pattern for SVM A is less clear, and we aim at investigating this in future work by conducting additional iterations.

%This phenomenon is in line with the goal of AL, i.e., creating a distance between positive and negative examples by focusing on borderline cases.

\textbf{Self-training} The rightmost part of Fig.~\ref{fig:beanUnlabelled} also illustrates how we evaluated self-training using data annotated by B.
As depicted by the dashed horizontal line, we explored adding different fractions of the most confidently classified examples (cf. the white bars) to the training set, annotated with the label predicted by the classifier.
As a proof-of-concept, we report our results from adding the following unlabelled examples: 1) 5\% positive examples, 2) 50\% negative examples, and 3) 5\% positive examples, and 50\% negative examples.
These additions represent adding unlabelled examples farther away from the hyperplane than 1.76 on the positive side, and 0.88 on the negative side.

\begin{table}
\begin{center}
\caption{Results from self-training as a complement to B's training set. Baseline shows the results from iteration 16, other rows show differences in absolute values.}
\label{tab:self}
\begin{tabular}{ |l|c|c|c|c| } 
 \hline
 \textbf{Approach} & \textbf{Accuracy} & \textbf{Precision} & \textbf{Recall} & \textbf{$F_1$-score} \\ 
 \hline
 Baseline & 0.700 & 0.574 & 0.344 & 0.428 \\
 \hline
 1) +5\% pos. & +0.019 & +0.024 & +0.096 & +0.080 \\ 
 \hline
 2) +50\% neg. & +0.025 & +0.069 & +0.037 & +0.046 \\ 
 \hline
 1) and 2) & +0.043 & +0.103 & +0.067 & +0.079 \\
 \hline
\end{tabular}
\end{center}
\end{table}

Table~\ref{tab:self} shows our results, compared to the baseline provided by iteration 16 without any self-training.
Our results show that active learning combined with self-training can be used to improve an SVM classifier for Stack Overflow posts.
Both adding positive and negative examples from the unlabelled examples can improve classification accuracy.
We obtained the best results when adding both types of data, resulting in improvements from the baseline corresponding to +4.3\% accuracy, +10.3\% precision, +6.7\% recall, and +7.9\% $F_{1}$-score.

\textbf{Limitations} Finally, we briefly discuss two aspects of threats to validity.
First, we stress that we have populated Table~\ref{tab:self} by cherry-picking results from successful self-training runs.
Most of our trial runs with self-training generated similar or worse results.
Using an approach to semi-exhaustively evaluate different self-training settings, in total running about 50 experimental runs, Table~\ref{tab:self} shows the best results we obtained.
However, our work is not a case of publication bias as we aim only to exhibit \textit{the existence of a phenomenon}~\cite{hannay_role_2008} -- a beneficial application of self-training when text mining software repositories. 
Most self-training settings might deteriorate the accuracy, and a more systematic approach to parameter tuning~\cite{borg_tuner:_2016} would probably identify even better settings.

Second, the \textit{external validity}~\cite{runeson_case_2012} of our work is limited.
AL might be better suited for other software engineering text annotation tasks with less human interpretation.
It is probable that another set of annotators, guided by other annotation guidelines, would result in a different inter-rater reliability.
As highlighted by Settles~\cite{settles_theories_2011}, while evolving annotation criteria is often a practical reality when applying AL, changes is a violation of the basic stability assumption. 
We also cannot claim that self-training is beneficial to all types of text mining tasks in software engineering.
What we can say, however, is that for our particular task of classifying Stack Overflow posts related to performance of software components, self-training yielded improvements -- and that is enough to recommend further research.

\section{Conclusion and Implications for Future Text Mining} \label{sec:conc}
We explored using AL and an SVM classifier for Stack Overflow posts with two alternating annotators.
The primary lesson learned is that AL and text mining appears to be a difficult combination, at least for short texts such as Stack Overflow posts.
In contrast to image classification tasks\footnote{Please refer to Karen Zack's viral tweets, e.g., ``chihuahua or muffin?'': http://ow.ly/zpF1308F7kK}, human Stack Overflow annotators must interpret incomplete information  presented with limited context -- differences in annotations are inevitable.
However, we argue that awareness of this intrinsic challenge of AL can be used to complement a traditional annotation process, i.e., AL can be used to identify the borderline cases that are worthwhile to discuss.

Based on our experiences, we present two recommendations when using AL for text mining software repositories.
First, the \textit{annotation criteria must continuously evolve}, in parallel to the annotators' interpretation of them, in line with coding guidelines for qualitative research~\cite{runeson_case_2012}.
It is not enough to simply count the number of differing labels, instead qualitative analysis is needed to identify any potential systematic differences -- before it is too late.
Second, we suggest that AL settings with multiple annotators should be designed with \textit{partly overlapping iterations to enable early detection of discrepancies}. 
The size of the labelled training set would increase at a slower rate with overlapping iterations, thus this must be balanced against the value of better annotator alignment.
In future attempts with AL, we plan to initially design iterations with 25\% overlap, and then gradually decrease it to 5\% as consensus increases.  

Based on the second author's AL process, we evaluated complementing the training set using self-training.
Our results are promising, we show that adding both positive and negative examples to the training set can increase the classification accuracy.
In a semi-structured approach, we achieved improvements of 4.3\% accuracy and 7.9\% $F_1$-score.
We stress that our findings do not suggest that self-training generally is a good idea, rather our results constitute a proof-of-concept that self-training can be successfully combined with AL. 
Furthermore, we expect that further improvements from self-training would be possible, and plan to conduct systematic parameter optimization as the next step~\cite{borg_tuner:_2016}.

\section*{Acknowledgment}
The work is partially supported by a research grant for the ORION project (reference number 20140218) from The Knowledge Foundation in Sweden, the Wallenberg Autonomous Systems and Software Program (WASP), and the Industrial Excellence Center EASE - Embedded Applications Software Engineering\footnote{http://ease.cs.lth.se}.

\bibliographystyle{IEEEtran}
\bibliography{msr2017} 

\begin{thebibliography}{10}
\providecommand{\url}[1]{#1}
\csname url@samestyle\endcsname
\providecommand{\newblock}{\relax}
\providecommand{\bibinfo}[2]{#2}
\providecommand{\BIBentrySTDinterwordspacing}{\spaceskip=0pt\relax}
\providecommand{\BIBentryALTinterwordstretchfactor}{4}
\providecommand{\BIBentryALTinterwordspacing}{\spaceskip=\fontdimen2\font plus
\BIBentryALTinterwordstretchfactor\fontdimen3\font minus
  \fontdimen4\font\relax}
\providecommand{\BIBforeignlanguage}[2]{{%
\expandafter\ifx\csname l@#1\endcsname\relax
\typeout{** WARNING: IEEEtran.bst: No hyphenation pattern has been}%
\typeout{** loaded for the language `#1'. Using the pattern for}%
\typeout{** the default language instead.}%
\else
\language=\csname l@#1\endcsname
\fi
#2}}
\providecommand{\BIBdecl}{\relax}
\BIBdecl

\bibitem{pereira_unreasonable_2009}
F.~Pereira, P.~Norvig, and A.~Halevy, ``The {Unreasonable} {Effectiveness} of
  {Data},'' \emph{IEEE Intelligent Systems}, vol.~24, no.~2, pp. 8--12, 2009.

\bibitem{banko_scaling_2001}
M.~Banko and E.~Brill, ``Scaling to {Very} {Very} {Large} {Corpora} for
  {Natural} {Language} {Disambiguation},'' in \emph{Proc. of the 39th {Annual}
  {Meeting} on {Association} for {Computational} {Linguistics}}, 2001, pp.
  26--33.

\bibitem{mitra_comparing_2015}
T.~Mitra, C.~Hutto, and E.~Gilbert, ``Comparing {Person}- and {Process}-centric
  {Strategies} for {Obtaining} {Quality} {Data} on {Amazon} {Mechanical}
  {Turk},'' in \emph{Proc. of the 33rd {Annual} {ACM} {Conference} on {Human}
  {Factors} in {Computing} {Systems}}, 2015, pp. 1345--1354.

\bibitem{sun_carial:_2012}
B.~Sun, G.~Shu, A.~Podgurski, and S.~Ray, ``{CARIAL}: {Cost}-{Aware} {Software}
  {Reliability} {Improvement} with {Active} {Learning},'' in \emph{Proc. of the
  5th {International} {Conference} on {Software} {Testing}, {Verification} and
  {Validation}}, 2012, pp. 360--369.

\bibitem{lu_adaptive_2012}
H.~Lu and B.~Cukic, ``An {Adaptive} {Approach} with {Active} {Learning} in
  {Software} {Fault} {Prediction},'' in \emph{Proc. of the 8th {International}
  {Conference} on {Predictive} {Models} in {Software} {Engineering}}, 2012, pp.
  79--88.

\bibitem{lu_defect_2014}
H.~Lu, E.~Kocaguneli, and B.~Cukic, ``Defect {Prediction} between {Software}
  {Versions} with {Active} {Learning} and {Dimensionality} {Reduction},'' in
  \emph{Proc. of the 25th {International} {Symposium} on {Software}
  {Reliability} {Engineering}}, 2014, pp. 312--322.

\bibitem{settles_active_2010}
B.~Settles, ``Active {Learning} {Literature} {Survey},'' University of
  Wisconsin-Madison, Tech. Rep. Computer Sciences Technical Report 1648, 2010.

\bibitem{lin_combining_2010}
Y.~Lin, C.~Sun, W.~Xiaolong, and W.~Xuan, ``Combining {Self} {Learning} and
  {Active} {Learning} for {Chinese} {Named} {Entity} {Recognition},''
  \emph{Journal of Software}, vol.~5, no.~5, pp. 530--537, 2010.

\bibitem{zhang_active_2016}
Z.~Zhang, E.~Pasolli, M.~Crawford, and J.~C. Tilton, ``An {Active} {Learning}
  {Framework} for {Hyperspectral} {Image} {Classification} {Using}
  {Hierarchical} {Segmentation},'' \emph{IEEE Journal of Selected Topics in
  Applied Earth Observations and Remote Sensing}, vol.~9, no.~2, pp. 640--654,
  2016.

\bibitem{richarz_annotating_2012}
J.~Richarz, S.~Vajda, and G.~Fink, ``Annotating {Handwritten} {Characters} with
  {Minimal} {Human} {Involvement} in a {Semi}-supervised {Learning}
  {Strategy},'' in \emph{Proc. of the {International} {Conference} on
  {Frontiers} in {Handwriting} {Recognition}}, 2012, pp. 23--28.

\bibitem{wohlin_supporting_2016}
C.~Wohlin, K.~Wnuk, D.~Smite, U.~Franke, D.~Badampudi, and A.~Cicchetti,
  ``\BIBforeignlanguage{en}{Supporting {Strategic} {Decision}-{Making} for
  {Selection} of {Software} {Assets}},'' in
  \emph{\BIBforeignlanguage{en}{Software {Business}}}, ser. Lecture {Notes} in
  {Business} {Information} {Processing}, A.~Maglyas and A.~Lamprecht,
  Eds.\hskip 1em plus 0.5em minus 0.4em\relax Springer, Cham, 2016, no. 240,
  pp. 1--15.

\bibitem{cicchetti_towards_2016}
A.~Cicchetti, M.~Borg, S.~Sentilles, K.~Wnuk, J.~Carlsson, and
  E.~Papatheocharous, ``Towards {Software} {Assets} {Origin} {Selection}
  {Supported} by a {Knowledge} {Repository},'' in \emph{Proc. of the 1st
  {International} {Workshop} on {Decision} {Making} in {Software}
  {Architecture}}, 2016.

\bibitem{barua_what_2012}
A.~Barua, S.~Thomas, and A.~Hassan, ``What are {Developers} {Talking} {About}?
  {An} {Analysis} of {Topics} and {Trends} in {Stack} {Overflow},''
  \emph{Empirical Software Engineering}, vol.~19, no.~3, pp. 619--654, 2012.

\bibitem{treude_how_2011}
C.~Treude, O.~Barzilay, and M.~Storey, ``How do {Programmers} {Ask} and
  {Answer} {Questions} on the {Web}?: {NIER} track,'' in \emph{Proc. of the
  33rd {International} {Conference} on {Software} {Engineering}}, 2011, pp.
  804--807.

\bibitem{soliman_architectural_2016}
M.~Soliman, M.~Galster, A.~Salama, and M.~Riebisch, ``Architectural {Knowledge}
  for {Technology} {Decisions} in {Developer} {Communities}: {An} {Exploratory}
  {Study} with {StackOverflow},'' in \emph{Proc. of the 13th {Working}
  {IEEE}/{IFIP} {Conference} on {Software} {Architecture}}, 2016, pp. 128--133.

\bibitem{allamanis_why_2013}
M.~Allamanis and C.~Sutton, ``Why, {When}, and {What}: {Analyzing} {Stack}
  {Overflow} {Questions} by {Topic}, {Type}, and {Code},'' in \emph{Proc. of
  the 10th {Working} {Conference} on {Mining} {Software} {Repositories}}, 2013,
  pp. 53--56.

\bibitem{ponzanelli_mining_2014}
L.~Ponzanelli, G.~Bavota, M.~Di~Penta, R.~Oliveto, and M.~Lanza, ``Mining
  {StackOverflow} to {Turn} the {IDE} into a {Self}-confident {Programming}
  {Prompter},'' in \emph{Proc. of the 11th {Working} {Conference} on {Mining}
  {Software} {Repositories}}, 2014, pp. 102--111.

\bibitem{chowdhury_mining_2015}
S.~Chowdhury and A.~Hindle, ``Mining {StackOverflow} to {Filter} out
  {Off}-topic {IRC} {Discussion},'' in \emph{Proc. of the 12th {Working}
  {Conference} on {Mining} {Software} {Repositories}}, 2015, pp. 422--425.

\bibitem{hassan_software_2010}
A.~Hassan and T.~Xie, ``Software {Intelligence}: {The} {Future} of {Mining}
  {Software} {Engineering} {Data},'' in \emph{Proc. of the {FSE}/{SDP}
  {Workshop} on {Future} of {Software} {Engineering} {Research}}, 2010, pp.
  161--166.

\bibitem{menzies_inductive_2011}
T.~Menzies, C.~Bird, T.~Zimmermann, W.~Schulte, and E.~Kocaganeli, ``The
  {Inductive} {Software} {Engineering} {Manifesto}: {Principles} for
  {Industrial} {Data} {Mining},'' in \emph{Proc. of the {International}
  {Workshop} on {Machine} {Learning} {Technologies} in {Software}
  {Engineering}}, 2011, pp. 19--26.

\bibitem{yarowsky_unsupervised_1995}
D.~Yarowsky, ``Unsupervised {Word} {Sense} {Disambiguation} {Rivaling}
  {Supervised} {Methods},'' in \emph{Proc. of the 33rd {Annual} {Meeting} on
  {Association} for {Computational} {Linguistics}}, 1995, pp. 189--196.

\bibitem{ying_mining_2015}
A.~Ying, ``Mining {Challenge} 2015: {Comparing} and {Combining} {Different}
  {Information} {Sources} on the {Stack} {Overflow} {Data} {Set},'' in
  \emph{Proc. of the 12th {Working} {Conference} on {Mining} {Software}
  {Repositories}}, 2015.

\bibitem{pedregosa_scikit-learn:_2011}
F.~Pedregosa, G.~Varoquaux, A.~Gramfort, V.~Michel, B.~Thirion, O.~Grisel,
  M.~Blondel, P.~Prettenhofer, R.~Weiss, and V.~Dubourg, ``Scikit-learn:
  {Machine} learning in {Python},'' \emph{Journal of Machine Learning
  Research}, vol.~12, no. Oct, pp. 2825--2830, 2011.

\bibitem{vapnik_nature_2000}
V.~Vapnik, \emph{The {Nature} of {Statistical} {Learning} {Theory}}.\hskip 1em
  plus 0.5em minus 0.4em\relax Springer New York, 2000.

\bibitem{settles_theories_2011}
B.~Settles, ``From {Theories} to {Queries}: {Active} {Learning} in
  {Practice},'' in \emph{Proc. of the {JMRL} {Workshop} on {Active} {Learning}
  and {Experimental} {Design}}, 2011.

\bibitem{feng_mistakes_2015}
G.~Feng, ``Mistakes and {How} to {Avoid} {Mistakes} in {Using} {Intercoder}
  {Reliability} {Indices},'' \emph{Methodology: European Journal of Research
  Methods for the Behavioral and Social Sciences}, vol.~11, no.~1, pp. 13--22,
  2015.

\bibitem{hannay_role_2008}
J.~Hannay and M.~Jorgensen, ``The {Role} of {Deliberate} {Artificial} {Design}
  {Elements} in {Software} {Engineering} {Experiments},'' \emph{IEEE
  Transactions on Software Engineering}, vol.~34, no.~2, pp. 242--259, 2008.

\bibitem{borg_tuner:_2016}
M.~Borg, ``{TuneR}: {A} {Framework} for {Tuning} {Software} {Engineering}
  {Tools} with {Hands}-{On} {Instructions} in {R},'' \emph{Journal of Software:
  Evolution and Process}, vol.~28, no.~6, pp. 427--459, 2016.

\bibitem{runeson_case_2012}
P.~Runeson, M.~Host, A.~Rainer, and B.~Regnell, \emph{Case {Study} {Research}
  in {Software} {Engineering}. {Guidelines} and {Examples}}.\hskip 1em plus
  0.5em minus 0.4em\relax Wiley, 2012.

\end{thebibliography}

\end{document}